\documentclass{article}
\usepackage{spconf,amsmath,graphicx}

\usepackage{framed,multirow}
\usepackage{amssymb}
\usepackage{latexsym}
\usepackage{lipsum}
\usepackage{url,cite}
\usepackage[table]{xcolor}
\usepackage{graphicx}
\usepackage[tight,footnotesize]{subfigure}
\usepackage{dsfont}
\usepackage{tabularx, tabulary, multirow, bigstrut}
\usepackage{booktabs}
\usepackage[american]{babel}
\usepackage[vlined]{algorithm2e} 
\usepackage{fixltx2e}
\usepackage[pagebackref=false,breaklinks=true,letterpaper=true,colorlinks,bookmarks=false]{hyperref}
\graphicspath{ {./figs/} }

\title{Information-Maximizing Sampling to Promote Tracking-by-Detection}
\name{Kourosh Meshgi$^{\star}$ \qquad Maryam Sadat Mirzaei$^{\dagger}$ \qquad Shigeyuki Oba$^{\star}$\thanks{This article is based on results obtained from a project commissioned by the NEDO and was supported by Post-K application development for exploratory challenges from Japan's MEXT.}}
\address{$^{\star}$ Graduate School of Informatics, Kyoto University, Japan \\
    $^{\dagger}$ RIKEN Center for Advanced Intelligence Project (AIP), Japan}

\begin{document}
\setlength{\abovedisplayskip}{3pt}
\setlength{\belowdisplayskip}{3pt}
%
\maketitle
\begin{abstract}

The performance of an adaptive tracking-by-detection algorithm not only depends on the classification and updating processes but also on the sampling. Typically, such trackers select their samples from the vicinity of the last predicted object location, or from its expected location using a pre-defined motion model, which does not exploit the contents of the samples nor the information provided by the classifier. We introduced the idea of most informative sampling, in which the sampler attempts to select samples that trouble the classifier of a discriminative tracker. We then proposed an active discriminative co-tracker that embed an adversarial sampler to increase its robustness against various tracking challenges. Experiments show that our proposed tracker outperforms state-of-the-art trackers on various benchmark videos.
\end{abstract}
\section{Introduction}
\label{sec1}

Visual Tracking is one of the most fundamental building blocks in understanding videos and motion in the real-world. 
%
Discriminative trackers formulate tracking as a foreground/background discrimination, to tackle problems of generative models such as complex non-linear dynamics of the object and background clutter \cite{tang2007co}. Correlation filters and tracking-by-detection are mainstreams of such trackers.  

\begin{figure}[!t]
\centering
\includegraphics[width=0.9\linewidth]{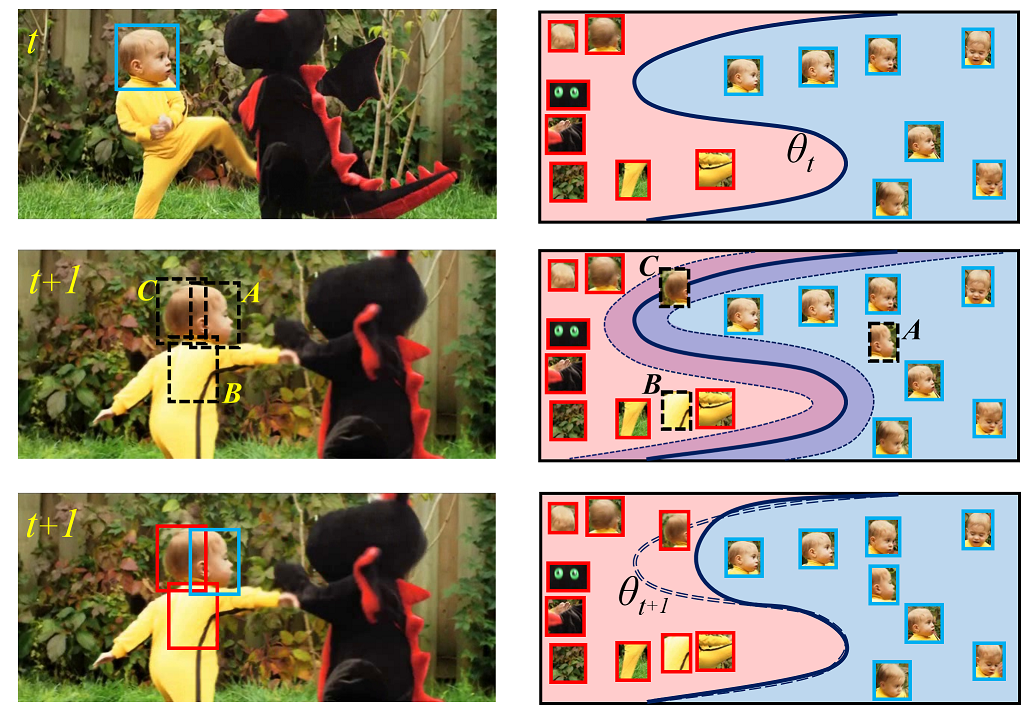}
\caption{In frame $t$ the sampler selects three samples \textit{A}, \textit{B}, and \textit{C} to evaluate by the classifier $\theta_t$. While samples \textit{A} and \textit{B} are easy for the classifier to classify, a label for sample \textit{C} would be uncertain. If $\theta_t$ is used to label sample \textit{C}, it would be misclassified as positive sample. On the other hand, since this sample is located near the decision boundary of classifier, knowing the correct label of \textit{C} is crucial to effectively update the model to $\theta_{t+1}$. Thus among these, \textit{C} is the most informative sample (principle of uncertain sampling) and an auxiliary classifier is needed to provide its label (co-learning). }
\label{fig:concept}
\vspace{-0.5 cm}
\end{figure}

Tracking-by-detection approaches \cite{kalal2012tracking,hare2011struck,zhang2014meem,hong2015multi,bertinetto2016staple,dinh2011context,tang2007co}, utilize one or more classifiers to classify the target. Despite their success in recent large benchmarks \cite{wu2013online,wu2015object}, these trackers still suffer from several shortcomings: \textit{(i) Uninformed sampling}, \textit{(ii) Label noise}: even the smallest mistakes in labeling are gradually accumulated in the self-learning loop of tracking-by-detection and cause a drift in the tracker, and \textit{(iii) Model drift}: an adaptive tracker should be updated rapidly yet remember the target appearance to recover from occlusions or target losses. Updating the model itself is not a straightforward task \cite{zhang2014meem}.
Label noise has been studied extensively, and various solutions such as robust loss functions \cite{masnadi2010design}, exploiting the information in unlabeled data \cite{grabner2008semi}, or even merging the sampling and learning process by structured learning is proposed \cite{hare2011struck}, ensemble-based trackers \cite{meshgi2016robust} and co-tracking \cite{tang2007co}. Model update, despite all the efforts, still challenges the performance of the trackers. Different online learning approaches (e.g., subspace, dictionary or incremental learning), as well as different update strategies (e.g., budgeted updating \cite{hare2011struck}, using auxiliary classifiers for sanity-check of the update  \cite{kalal2012tracking}, rolling back bad updates \cite{zhang2014meem}, and combining long and short-term memories \cite{hong2015multi}), tried to alleviate this problem.

\begin{figure*}
\centering
\includegraphics[width=0.75\linewidth]{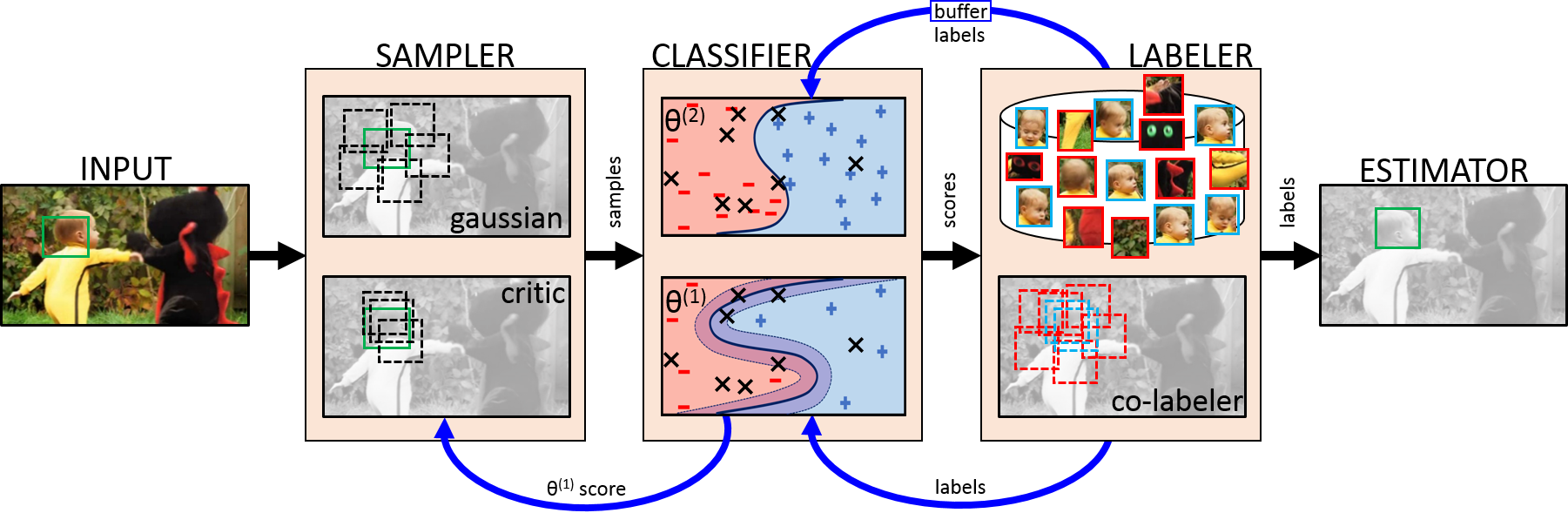}
\caption{Schematic of the system. The proposed tracker, collect half of the sample from a Gaussian distribution around the last target state. Meanwhile, the critic learns how to generate transformations that maximally challenge the short-memory classifier ($\theta^{(1)}$), in order to accelerate learning and improve the accuracy. The classifiers exchange information via an active learning scheme, to realize a collaborative robust labeling. The classifiers are then updated and the target state is estimated (See Alg. \ref{alg:asct}).}
\label{fig:schematic}
\vspace{-0.5 cm}
\end{figure*}

To define the region-of-interest for sampling, some tracking-by-detection used context information \cite{dinh2011context}, optical flow \cite{kalal2012tracking}, dynamics model \cite{cehovin2013robust}, or a pool of motion models \cite{kwon2011tracking} but most of these methods suffers from sudden failures. This is mainly characterized to the assumption that the last predicted target location is accurate, an assumption that can be violated under challenging real-world scenarios such as abrupt or fast target movements, occlusions or severe clutters. Object proposal generators, such as Edge boxes \cite{zitnick2014edge}, CPMC \cite{carreira2010constrained}, and Selective Search \cite{uijlings2013selective}, are a group of models that provide a fine-selected set of candidates that potentially contain the object in the image. These models --provided that they can provide efficient and reliable candidates-- can serve as the sampler in a tracking-by-detection pipeline, such as Edge Box which is employed in EBT \cite{zhu2016beyond}.

In this study, to illustrate the role of sampling in providing better information for the classifier, we address the problem of uninformed sampling by proposing information-maximizing sampling, in which a ``critic'' unit learns to sample the essential parts of the image: the most informative ones for the classifier. It is realized by using a semi-supervised co-tracking framework in which the information exchange is managed by active learning, selecting the most uncertain samples of each classifier and querying it from the other. 
In this framework, two classifiers are used: one with long-term memory that is updated infrequently, and the other with a short-term memory updated every frame. The ``critic'' in this framework learn to challenge the latter classifier by monitoring that classifier's performance over various samples to select future samples that challenge the classifier the most forcing it to collaborate with the long-memory classifier.
Using critic solves uninformed sampling, active learning tackles treat samples unequally, co-learning addresses label noise and together with short-long memory mixture resist model drift. The proposed tracker demonstrate a superior performance compared to the state-of-the-art on challenging sequences. 

\section{Proposed Method}

\subsection{Tracking by Detection}
Online visual tracking is the task of finding the proper transformation $\mathbf{y}_t$ that transform the previous state $\mathbf{p}_{t-1}$ into the new state $\mathbf{p}_t = \mathbf{p}_{t-1} \circ \mathbf{y}_t$. In tracking-by-detection framework, it is realized by obtaining several samples $\mathbf{x}^j_t (j=1,\ldots,n)$ from the new frame and evaluating them using a classifier $\theta_t$ to distinguish if they contain the target or the background. The most confident sample according to the classifier is typically selected as the next target state. To accommodate target evolutions throughout the tracking scenario, the classifier should be updated.

A typical pipeline for this process is to sample transformations $\mathbf{y}_t \in \mathcal{Y}_t$ using dense sampling or sparse sampling with regard on the previous target state, $\mathbf{p}^j_t = \mathbf{p}_{t-1} \circ \mathbf{y}^j_t \in \mathcal{P}_t$. The samples are defined using these transformations, and their corresponding image patches $\mathbf{x}^j_t \in \mathcal{X}_t$ is selected from image. After an optional feature extraction stage, these image patches are evaluated by classifier $\theta_t$ and scored by its scoring function $h: \mathcal{X} \rightarrow \mathbb{R}$.
\begin{equation}
s^j_t = h \big( \mathbf{x}_t^{\mathbf{p}_{t-1} \circ \mathbf{y}^j_t}| \theta_t \big).
\label{eq:score}
\end{equation}
If the score is above a threshold $\tau$, the sample is considered as a possible target match,
\begin{equation}
\ell^j_t = \mathrm{sign} ( s^j_t - \tau).
\label{eq:label}
\end{equation}
A weighted average of the positive samples is selected as the next target state (indicating its location and size),
\begin{equation}
\hat{\mathbf{y}}_t = \sum_k s_t^k \mathbf{y}^k_t \;,\mathrm{s.t.}\; \ell^k_t > 0 (j=1,\ldots,n).
\label{eq:estimate}
\end{equation}
Finally, the classifier is updated by its own labeled data,
\begin{equation}
\theta_{t+1} = u(\theta_t,\mathcal{X}_t,\mathcal{L}_t)
\label{eq:update}
\end{equation}
in which $u(.)$ is the update function (e.g., budgeted SVM update \cite{hare2011struck}), and $\ell^j_t \in \mathcal{L}_t$ is the corresponding labels of $\mathcal{X}_t$.

\subsection{Information Maximizing Sampling Strategy}
\label{sec:sampling}
Obtaining samples for a discriminative tracker, have been understudied in the literature. While dense sampling using sliding windows, Gaussian sampling around the last known target location ($\mathbf{y}^j_t \sim \mathcal{N}(\mathbf{p}_{t-1},\Sigma_{search})$), using 1st or 2nd-degree motion models, and particle filters are known as successful practices for sampling, still the need for an informed sampling that uses the content of samples to obtain better samples is needed. In addition, circular shift \cite{henriques2015high} to increase the number of positive samples, despite its computation efficiency benefits, inject noise into the tracking loop in the long-term that leads to tracking drift \cite{kiani2017learning}. To address this issue, the content of the samples must be considered in sample selection to realize an informed sampling. For instance, in \cite{zhu2016beyond} the silhouette of the target is searched within samples to provide fewer samples with a higher chance of being the target. However as argued in \cite{hare2011struck}, the sampling and classification have two different objectives. While the former tries to provide better samples from the target, the latter tries to construct a better classifier, demanding representative negative samples and supports for defining an accurate classification boundary. 

In this study, we take another approach, by exploiting the uncertainties of the classifier, we try to obtain samples that knowing their labels, would maximally improve the classification accuracy, in other words, \textit{most informative samples}. 
 
\begin{algorithm}[!t]
\DontPrintSemicolon
\SetKwInOut{Input}{input}\SetKwInOut{Output}{output}
\Input{Last state $\mathbf{p}_{t-1}$, Classifiers $\theta_t^{(i)}$, Critic $\Psi$}
\Output{New state $\mathbf{p}_t$, Updated models $\theta_{t+1}^{(i)},\Psi_{t+1}$}
\BlankLine
\For{$j \leftarrow 1$ \KwTo $n$ }
{
\uIf(){$j < \frac{n}{2}$}
        {
        \emph{Random sampling} $\mathbf{y}^j_t \sim \mathcal{N}(\mathbf{p}_{t-1},\Sigma_{search})$\;
        }
\Else
        {
        \emph{Guided sampling} $\mathbf{y}^j_t \sim g(\mathbf{p}_{t-1} |\Psi )$\;
        }
\emph{Calculate position and score} (eq\eqref{eq:score_asct})\;
\emph{Obtain label and queries} (eq(\ref{eq:label_asct},~\ref{eq:query_asct})\;
\emph{Calculate error rate and weights} (eq(\ref{eq:error_asct},~\ref{eq:weight_asct})\;
\emph{Update critic} (eq\eqref{eq:update_critic})\;
}
\emph{Update classifiers} (eq(\ref{eq:update_short},~\ref{eq:update_long})\;
\emph{Estimate the transformation and new state} (eq\eqref{eq:estimate})\;

\BlankLine
\caption{Information Maximizing Sampling Tracker}
\label{alg:asct}
\vspace{-0.1em}
\end{algorithm}
Recently, generative adversarial networks implemented by a system of two neural networks competing against each other in a zero-sum game framework \cite{goodfellow2014generative} gains much attention. In this framework, one network is generative which is taught to map from a latent space to a particular data distribution, and the other is a discriminative network that is simultaneously taught to discriminate between true data and synthesized instances produced by the generator. Inspired by this framework, we proposed a ``critic'' that tries to expose the weaknesses of the tracker's classifier, and the classifier tries to improve its classification in those area to provide a good classification for those sort of samples. To this end, we employed a structured learning for critic, with learning prediction function $g: \mathcal{X}_t \rightarrow \mathcal{Y}_t$. In our approach, a labeled example is a pair $\langle \mathbf{x}^{\mathbf{p}_{t-1}}, \mathbf{y}^j_t \rangle$ where $\mathbf{y}^j_t$ is a challenging transformation given the last known target position $\mathbf{p}_{t-1}$. We learn $G: \mathcal{X}_t \times \mathcal{Y}_t \rightarrow \mathbb{R}$ on-the-fly using a structured-output SVM framework governed by $\Psi$ which introduces a discriminant function  that can be used for prediction
\begin{equation}
\mathbf{y}^j_t = g(\mathbf{p}_{t-1} |\Psi ) = \mathrm{max}_{k=1,\ldots,j-1} G(\mathbf{p}_{t-1}|\mathbf{y}^k_t,\Psi)
\label{eq:critic}
\end{equation}
The critic is updated with every selected sample $\mathbf{x}^j_t$ and its label $\ell^j_t$ to help finding the next challenging sample,
\begin{equation}
\Psi \leftarrow u_c(\Psi,\mathbf{x}^j_t,\ell^j_t),
\label{eq:update_critic}
\end{equation}
where $u_c(.)$ is a budgeted SVM update inspired by \cite{hare2011struck}.
If the last generated sample falls within the uncertain region of the classifier, its addition to the critic reinforce the ability of the critic to generate challenging samples similar to the last sample, otherwise, it signals the critic to explore other ways of generating samples to challenge the main classifier.

\subsection{Information Maximizing Sampling Tracking}
The proposed tracker is consisted of two classifiers  $\theta_t^{(1)}$ and $\theta_t^{(2)}$, having short-term and long-term memory respectively. This mixture of memories balances the stability-plasticity of the tracker. The data exchange of two classifiers is conducted by active learning, in which the most uncertain samples of one classifier is labeled by the other classifier. Finally, these labeled data are used to update the classifiers. 

To realize an informed sampling, we proposed a hybrid of Gaussian sampling (based on the last known target position) and critic-generated sampling (that challenges the main classifier to improve its decision boundary). In each frame $t$, half of the samples are obtained using the Gaussian sampling, classified and use to update the critic with the recent changes of the target and background. Then the critic, finds several challenging samples (\textit{candidates}) for $\theta_t^{(1)}$ using eq\eqref{eq:critic}. If a candidate is not challenging for the classifier, $|h \big( \mathbf{x}_t^{\mathbf{p}_{t-1} \circ \mathbf{y}^j_t}| \theta_t^{(1)} \big)| \geq \tau_t$ , it is discarded and a new candidate is seek. The selected samples are scored using
\begin{equation}
s^{j,(i)}_t = h \big( \mathbf{x}_t^{\mathbf{p}_{t-1} \circ \mathbf{y}^j_t}| \theta_t^{(i)} \big)
\label{eq:score_asct}
\end{equation}
and their labels are determined by
\begin{align}
\ell^j_t &=
  \begin{cases}
   \mathrm{sign} (s^{j,(1)}_t)        &, s^{j,(2)}_t < \tau_t , s^{j,(1)}_t \geq \tau_t \\
   \mathrm{sign} (s^{j,(2)}_t)        &, s^{j,(1)}_t < \tau_t , s^{j,(2)}_t \geq \tau_t \\
   \mathrm{sign} (\alpha_t^{(1)}s^{j,(1)}_t + \alpha_t^{(2)}s^{j,(2)}_t)         &, \text{otherwise}
  \end{cases}
  \label{eq:label_asct}
\end{align}

Next, the queries of classifiers (i.e., the data they want the other classifier to label) are determined following the principle of uncertainty sampling \cite{lewis1994sequential}
\begin{equation}
q^{1 \rightarrow 2}_t = \{ \mathbf{x}_t^j | s^{j,(1)}_t < \tau_t \}.
\label{eq:query_asct}
\end{equation}
Here, $\tau_t$ is selected such that the $m$ most uncertain samples falls in $q^{1 \rightarrow 2}_t$.
To calculate the weight of classifiers, first their errors are calculated as the number of mismatches between the classifier label and the co-tracking label, 
\begin{equation}
e^{(i)}_t = \sum_j \mathds{1} (\ell^j_t \neq \mathrm{sign} (s^{j,(i)}_t)),
\label{eq:error_asct}
\end{equation}
then it is used to calculate the weights of classifiers
\begin{equation}
\alpha^{(i)}_t = 1-\frac{e^{(i)}_t + \epsilon}{\sum_{i \in \{1,2\} } e^{(i)}_t + \epsilon},
\label{eq:weight_asct}
\end{equation}
where $\mathds{1}(.)$ is the indicator function and $\epsilon$ is a small constant. 
After having all samples update short-term classifier
\begin{equation}
\theta^{(1)}_{t+1} = u_1(\theta_t^{(1)},q_t^{2 \rightarrow 1},\mathcal{L}_t)
\label{eq:update_short}
\end{equation}
And long-term one
\begin{align}
\theta^{(2)}_{t+1} &=
\begin{cases}
 u_2(\theta^{(2)}_t,\mathcal{X}_{t-\Delta,\ldots,t},\mathcal{L}_{t-\Delta,\ldots,t}) &, t=k\Delta \\
 \theta^{(2)}_t  &, t \neq k\Delta 
\end{cases}
\label{eq:update_long}
\end{align}
Then the target transformation is estimated from eq\eqref{eq:estimate} and determine the new state from $\mathbf{p}_{t} = \mathbf{p}_{t-1} \circ \mathbf{y}_t$. Algorithm \ref{alg:asct} summarizes the proposed tracker. Tracker's parameters ($n$, $\Sigma_{\mathrm{search}}$, $m$ and $\Delta$) are determined with the cross-validation.

\section{Evaluation}
\label{sec3}
\begin{figure}[!t]   
\centering
\includegraphics[width= 0.49\linewidth]{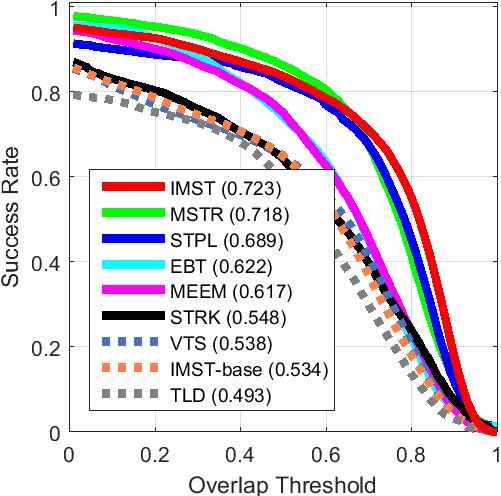}
\includegraphics[width= 0.49\linewidth]{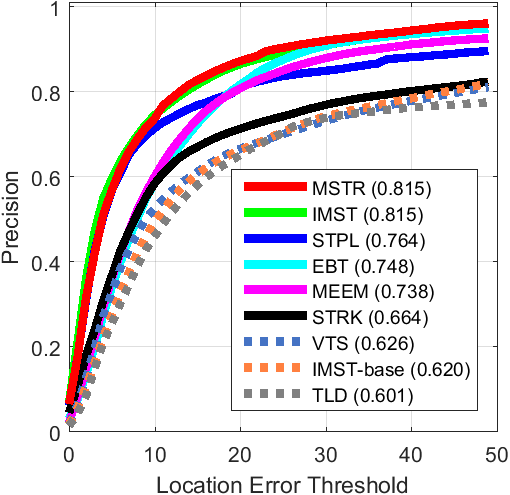}
\caption{Quantitative evaluation of trackers using precision plot (left) and success plot (right) for all sequences in OTB-50\cite{wu2013online}. The AUC of plots are used for fair comparison.}
\label{fig:precision}
\vspace{-0.3 cm}
\end{figure}

\vspace*{-\baselineskip}
\begin{table}[!b]
\caption{Quantitative evaluation of trackers under different tracking challenges using AUC(\%) of success plot on OTB-50. The {\color{red}first}, {\color{green}second} and {\color{blue}third} best results are shown in color.}
\label{tab:attributes}
\centering
\scalebox{0.85}{
\renewcommand{\arraystretch}{0.9}
\begin{tabular}{@{}>{\bfseries}l c c c c c c c c@{}} \toprule
\makebox[9mm]{Attribute} & \makebox[7mm]{VTS} &\makebox[7mm]{TLD} &\makebox[7mm]{STRK} &\makebox[7mm]{MEEM} &\makebox[7mm]{STPL} &\makebox[7mm]{MSTR} &\makebox[7mm]{EBT} & \makebox[7mm]{IMST}\\ \midrule
    IV  & {\color{black}54.3} & {\color{black}47.8} & {\color{black}53.0} & {\color{black}62.3} & {\color{blue} 67.7} & {\color{red}  72.6} & {\color{black}61.2} & {\color{red}  72.6} \\ 
    SV  & {\color{black}51.8} & {\color{black}49.1} & {\color{black}51.8} & {\color{black}58.3} & {\color{blue} 67.6} & {\color{green}70.6} & {\color{black}58.0} & {\color{red}  72.4} \\  
    IPR & {\color{black}55.1} & {\color{black}50.4} & {\color{black}53.7} & {\color{black}57.7} & {\color{green}68.9} & {\color{blue} 68.5} & {\color{black}56.9} & {\color{red}  73.5} \\  
    OPR & {\color{black}54.9} & {\color{black}47.8} & {\color{black}53.2} & {\color{black}62.1} & {\color{blue} 67.5} & {\color{green}70.2} & {\color{black}61.4} & {\color{red}  72.5} \\  
    DEF & {\color{black}54.1} & {\color{black}38.2} & {\color{black}51.3} & {\color{black}61.9} & {\color{red}  70.4} & {\color{green}68.9} & {\color{black}64.6} & {\color{blue} 66.1} \\  
    OCC & {\color{black}52.2} & {\color{black}46.1} & {\color{black}50.2} & {\color{black}60.8} & {\color{blue} 69.1} & {\color{green}71.0} & {\color{black}59.6} & {\color{red}  71.8} \\  
    OV  & {\color{black}51.5} & {\color{black}53.5} & {\color{black}51.5} & {\color{black}68.5} & {\color{black}61.8} & {\color{red}  73.3} & {\color{blue} 70.0} & {\color{green}71.1} \\  
    LR  & {\color{black}35.0} & {\color{black}36.2} & {\color{black}33.3} & {\color{black}43.5} & {\color{blue} 47.4} & {\color{green}50.2} & {\color{black}30.6} & {\color{red}  56.3} \\  
    BC  & {\color{black}56.7} & {\color{black}39.4} & {\color{black}51.5} & {\color{black}67.0} & {\color{black}66.9} & {\color{red}  71.7} & {\color{blue} 67.2} & {\color{green}71.2} \\  
    FM  & {\color{black}43.5} & {\color{black}44.6} & {\color{black}52.0} & {\color{blue} 64.6} & {\color{black}55.9} & {\color{green}65.0} & {\color{red}  65.4} & {\color{black}64.3} \\  
    MB  & {\color{black}39.4} & {\color{black}41.0} & {\color{black}46.7} & {\color{black}62.8} & {\color{black}61.5} & {\color{green}65.2} & {\color{blue} 63.8} & {\color{red}  65.5} \\ \midrule
    FPS & {\color{black} 5.3} & {\color{blue} 21.2} & {\color{black}11.3} & {\color{black}14.2} & {\color{red}  48.1} & {\color{black} 8.3} & {\color{black} 3.8} & {\color{green}23.2} \\ \bottomrule
\end{tabular}
}
\vspace{-0.5cm}
\end{table}
In this section, we compare the proposed tracker, with its baseline (the co-tracker with a long and a short memory) and the state-of-the-art algorithms on the object tracking benchmark (OTB-50 \cite{wu2013online}). The sequences of OTB-50 are attributed by one or more tracking challenges: illumination (IV), scale (SV), in-plane rotation (IPR), out-of-plane rotation (OPR), deformation (DEF), occlusion (OCC), out-of-view (OV), background clutter (BC), low resolution (LR), fast motion (FM) and motion blur (MB). The performance of the trackers is compared with the area under curve of success plot and precision plot, on all of the sequences (Figure \ref{fig:precision}), or a subset of them with the given attribute (Table \ref{tab:attributes}). Success plot indicates the reliability of the tracker and its overall performance while precision plot reflects the accuracy of the localization. 
Figure \ref{fig:precision} presents that using the proposed sampling method by keeping a fixed number of samples significantly improved the performance of the IMST tracker, over its baseline.
To establish a fair comparison with the state-of-the-art of tracking-by-detection algorithms, TLD \cite{kalal2012tracking} and STRUCK \cite{hare2011struck} are selected based on the results of \cite{wu2013online}, MUSTer \cite{hong2015multi}, STAPLE \cite{bertinetto2016staple} and MEEM \cite{zhang2014meem} are selected based on the results of VOT2016, and VTS \cite{kwon2011tracking} and EBT \cite{zhu2016beyond} was selected to compare the effectiveness of sampling methods. The results reported here is the average of five independent runs.
As Figure \ref{fig:precision} and Table \ref{tab:attributes} illustrates, the proposed tracker outperforms the state-of-the-art. The proposed algorithm also has superior performance in most of the subcategories. High performance in SV, LR, OV, and MB are specifically the results of proposed sampling and comparable results in BC by adding the short/long memory combination.

\begin{figure}[!t]
\centering
\includegraphics[width= 0.32\linewidth]{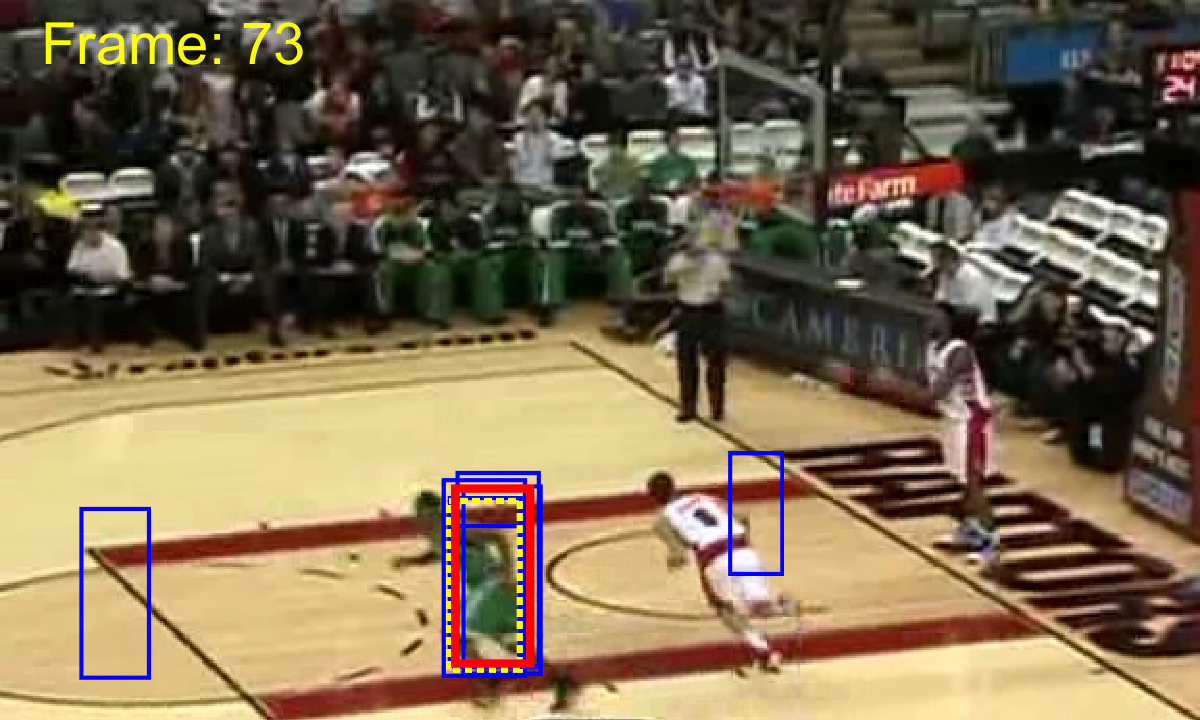}
\includegraphics[width= 0.32\linewidth]{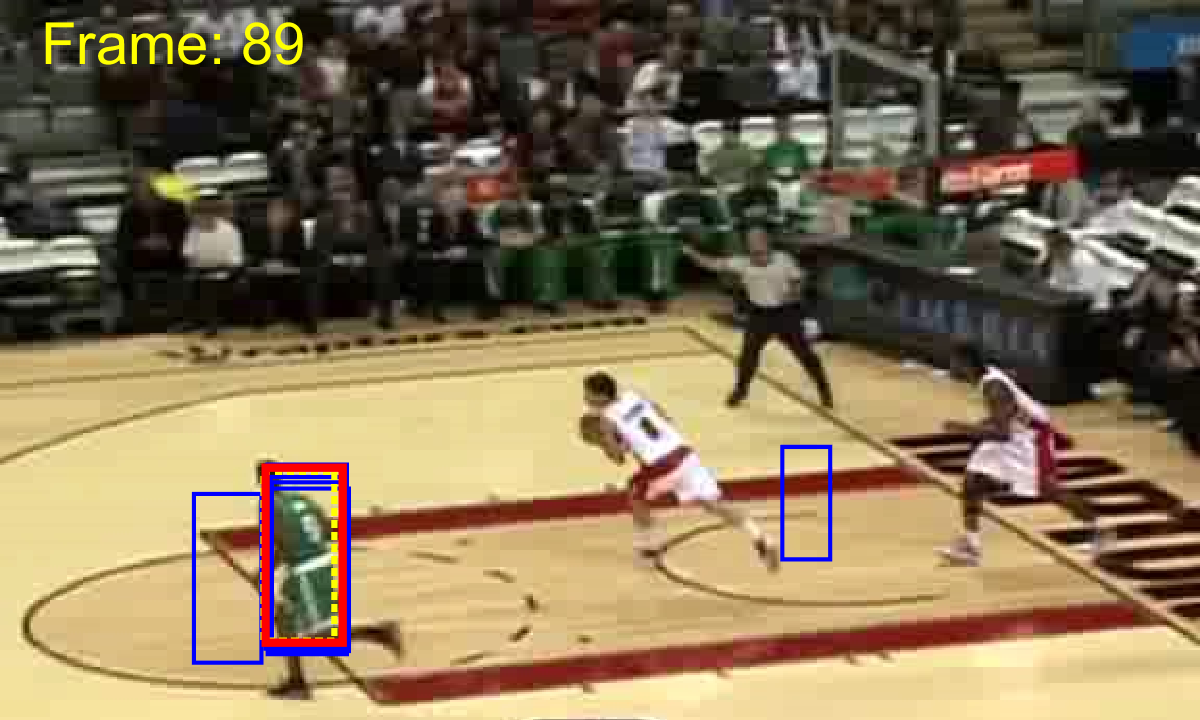}
\includegraphics[width= 0.32\linewidth]{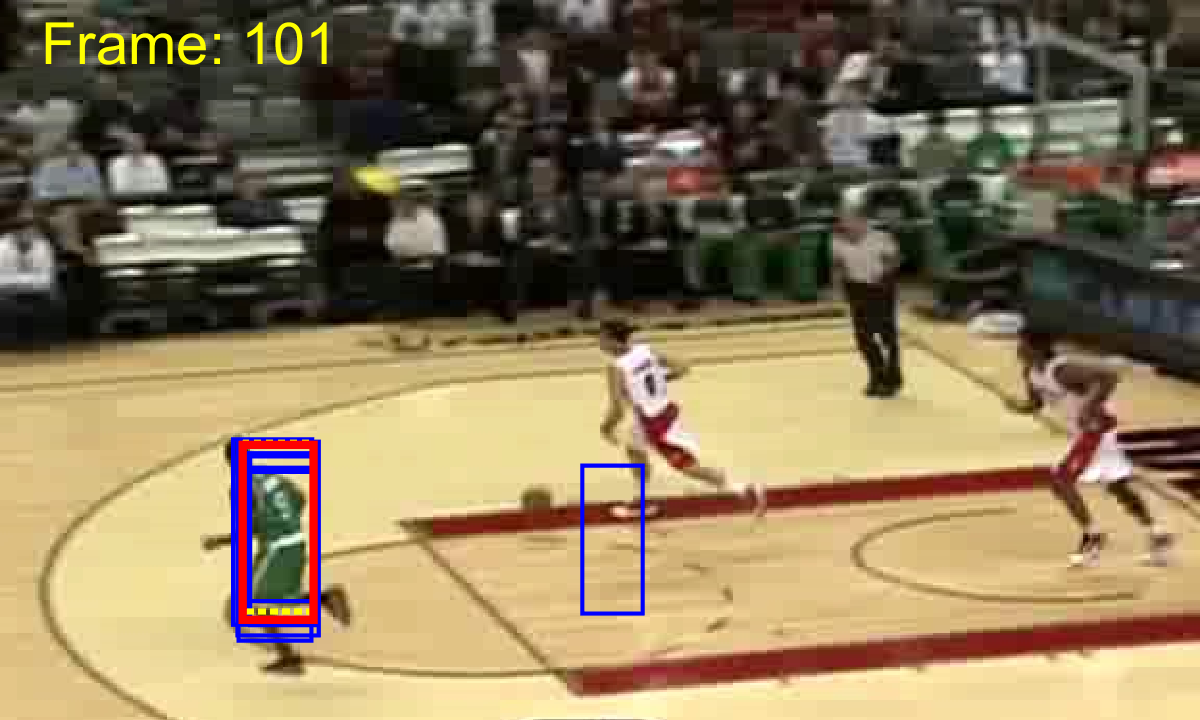}
\includegraphics[width= 0.32\linewidth]{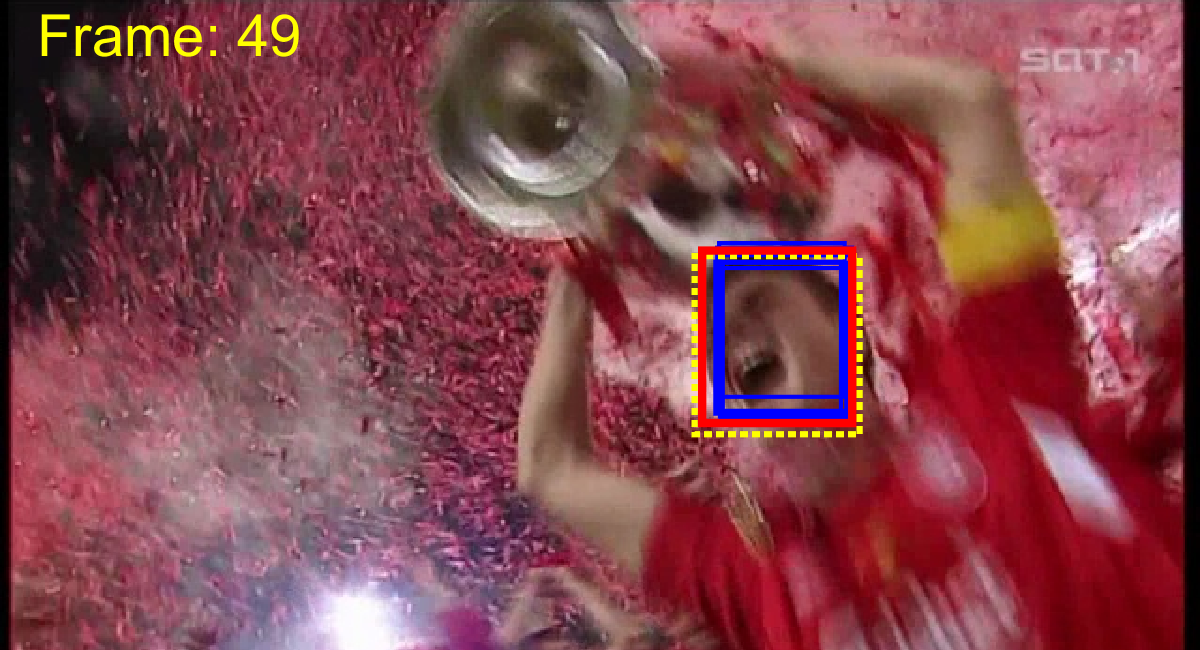}
\includegraphics[width= 0.32\linewidth]{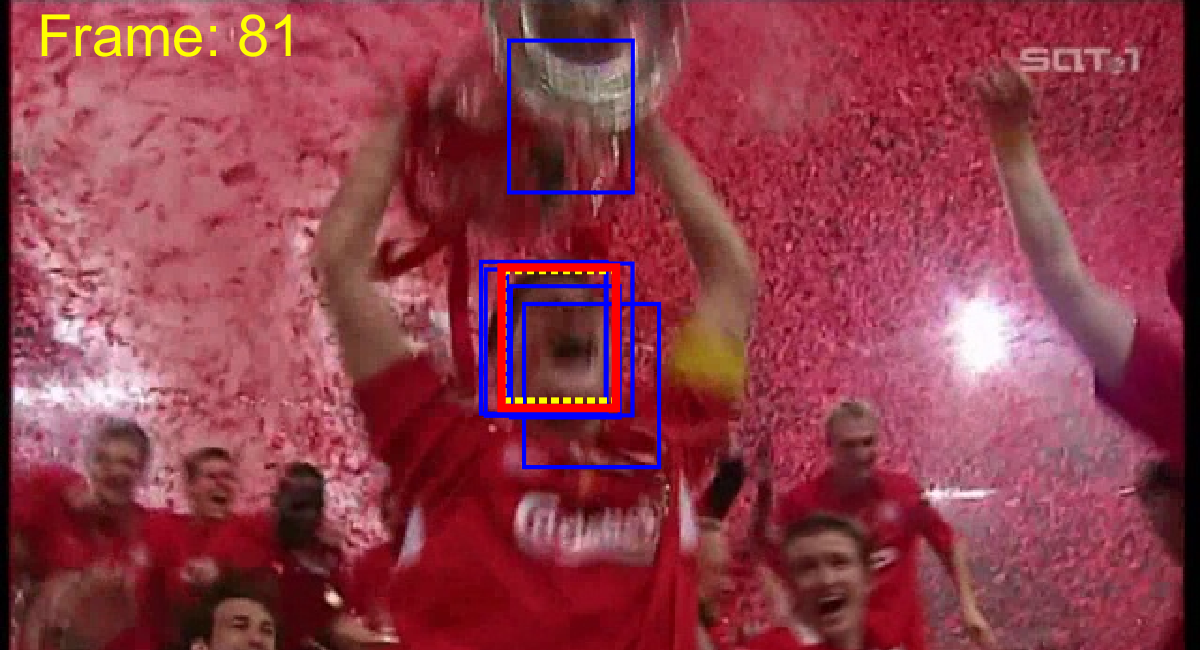}
\includegraphics[width= 0.32\linewidth]{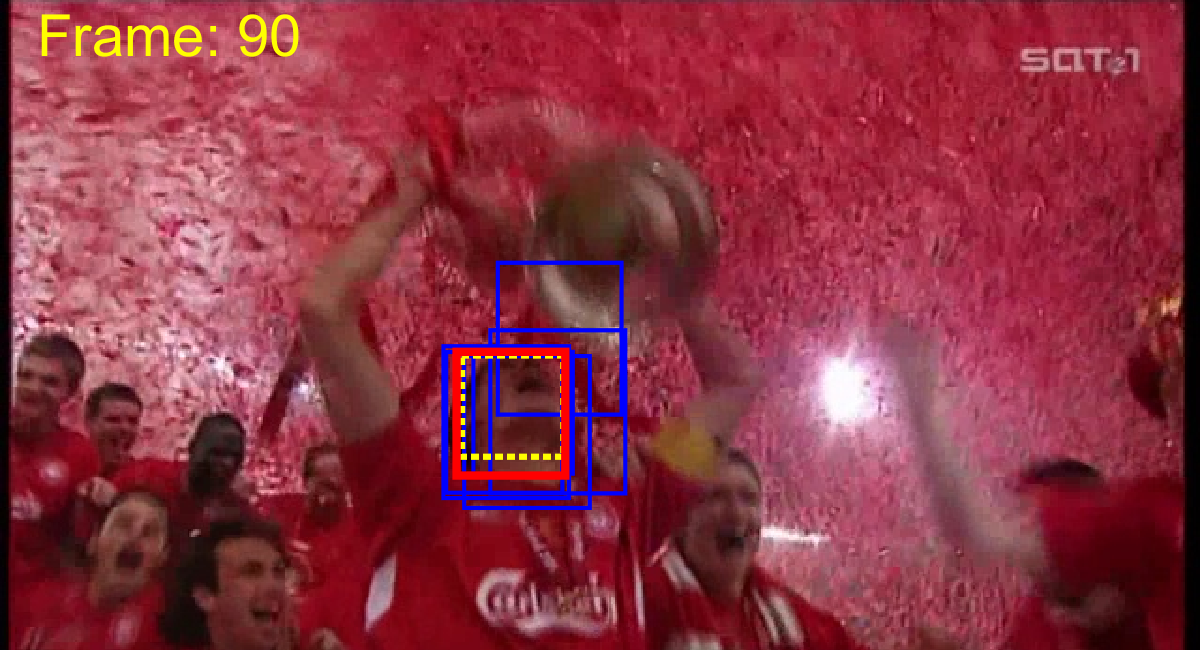}
\includegraphics[width= 0.32\linewidth]{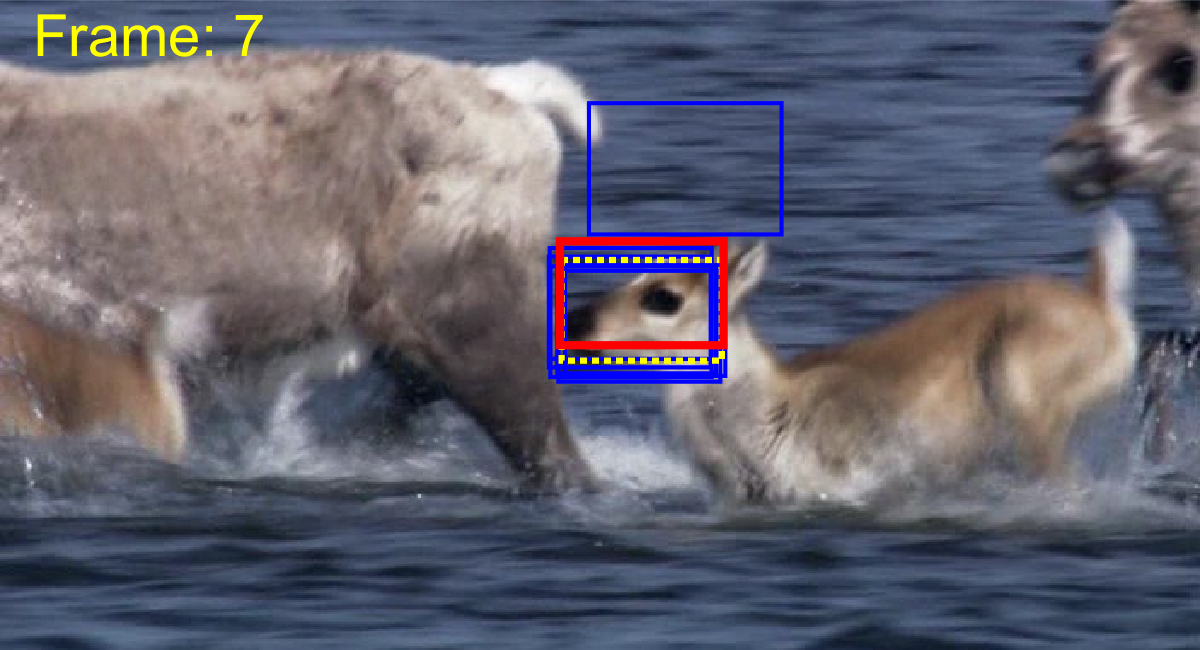}
\includegraphics[width= 0.32\linewidth]{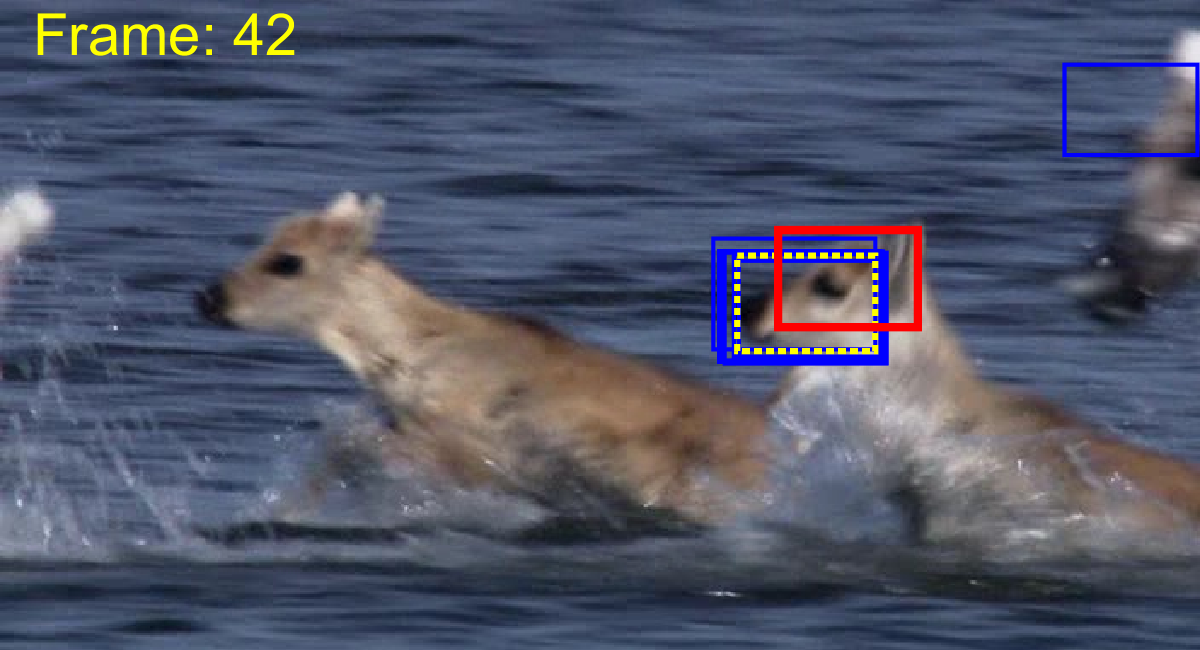}
\includegraphics[width= 0.32\linewidth]{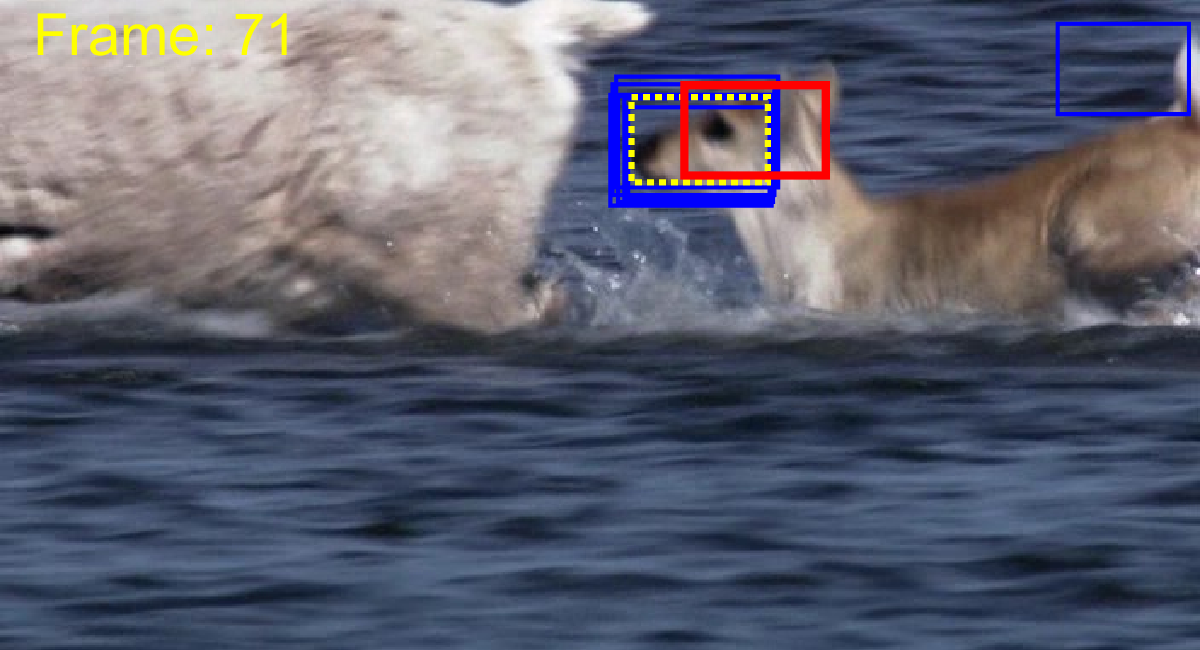}
\includegraphics[width= 0.32\linewidth]{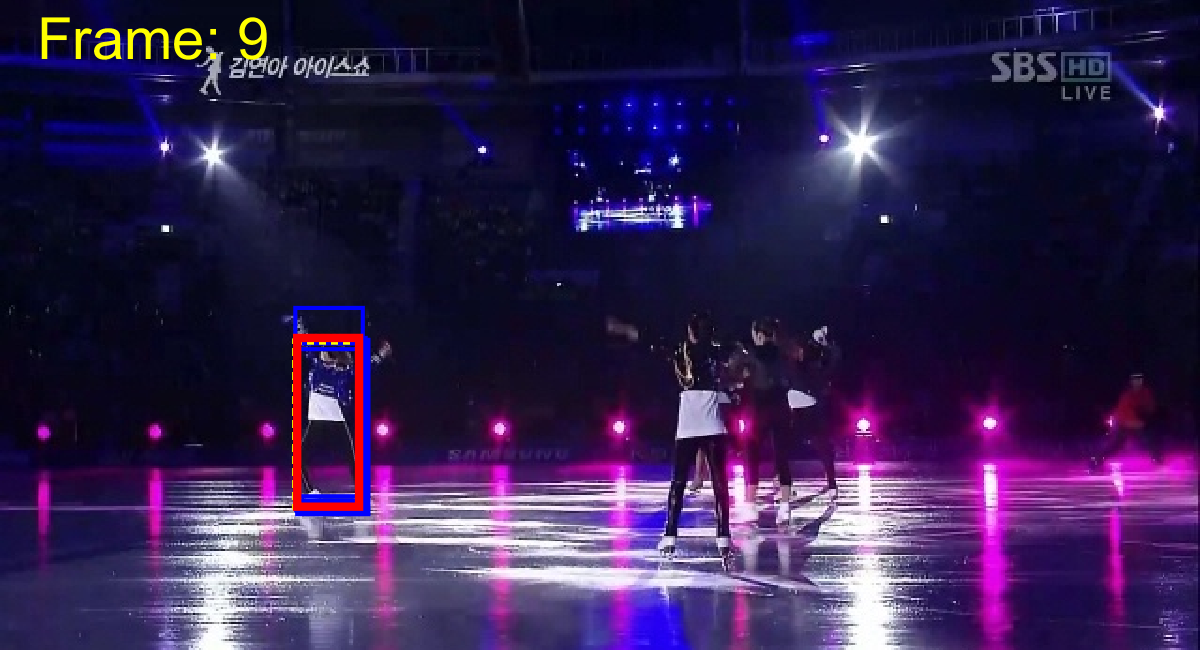}
\includegraphics[width= 0.32\linewidth]{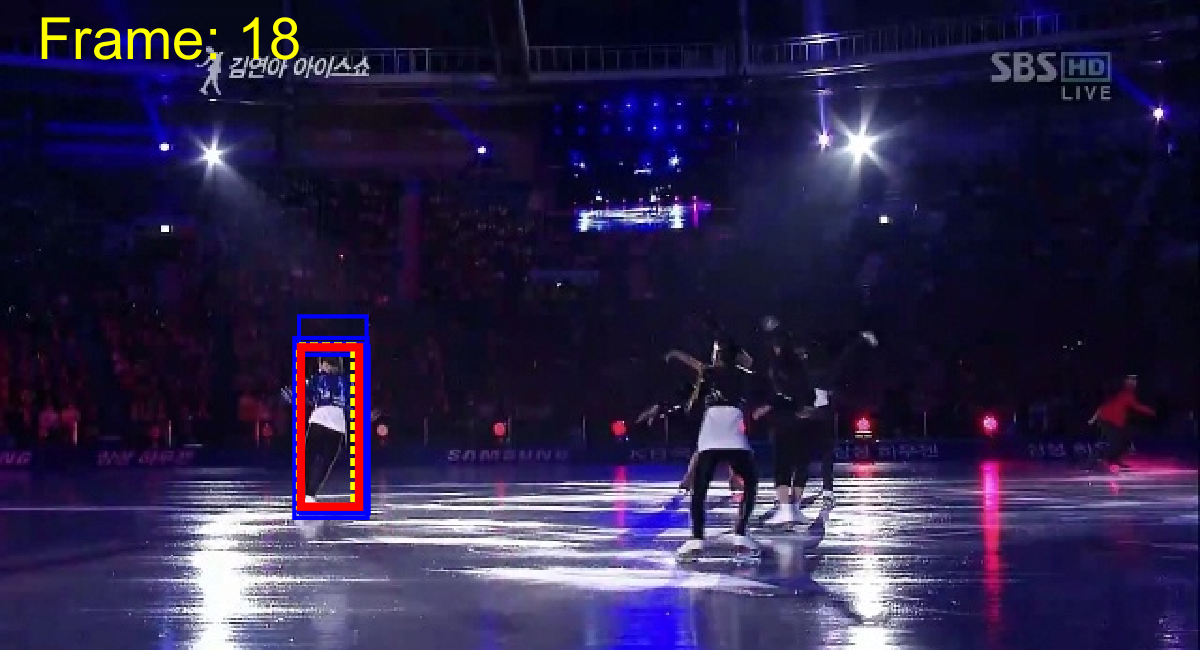}
\includegraphics[width= 0.32\linewidth]{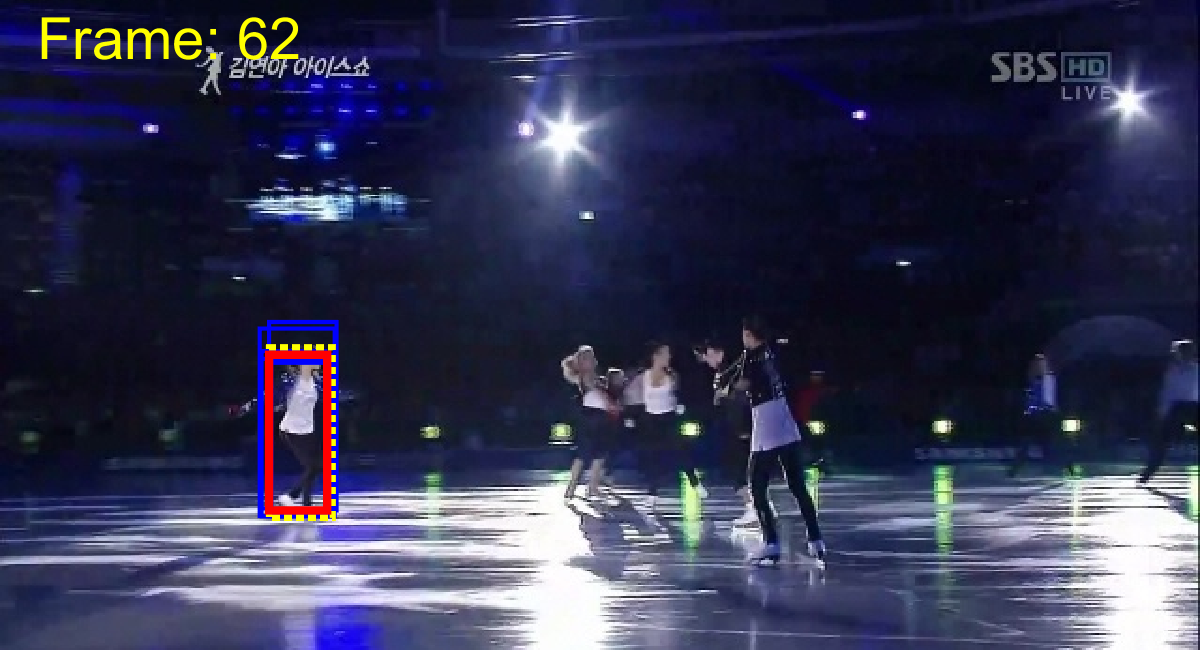}
\caption{Exemplary tracking results of proposed tracker (in red) and other evaluated trackers (blue) on several challenging video sequences. The ground truth is depicted in yellow. More results are available from \url{http://ishiilab.jp/member/meshgi-k/imst.html}.
}
\label{fig:eval_qual}
\vspace{-0.5cm}
\end{figure}

\section{Conclusion}
\label{sec:conclusion}
In this study, we have proposed an information maximizing sampling paradigm to be integrated into a discriminative active co-tracker. It is realized by a structured learning scheme which maps the sample space to transformation space and select the most informative samples to accelerate classifier learning and foster the accurate tracking. The proposed tracker, IMST, obtain samples by considering target's spatiotemporal properties and uncertainty-analysis of the classifier and provides the required labels from a long-memory auxiliary classifier and outperformed the state-of-the-art. 


\vfill
\pagebreak


\bibliographystyle{IEEEbib}
\def\IEEEbibitemsep{-2pt plus .5pt}
\bibliography{refs}

\end{document}